\documentclass{article} 
\usepackage{colm2024_conference}

\usepackage{microtype}
\usepackage{hyperref}
\usepackage{url}
\usepackage{booktabs}
\usepackage{amsfonts}
\usepackage{nicefrac}
\usepackage{xcolor}
\usepackage{inconsolata}
\usepackage{graphicx}
\usepackage{subfigure}
\usepackage{amsmath}

\usepackage{geometry}
\usepackage{tabularray}
\usepackage{caption}
\usepackage{lipsum}

\usepackage{array}
\usepackage{longtable}
\usepackage{booktabs}
\usepackage{makecell}

\title{Instruction-Driven Game Engines on Large Language Models}

\author{
  Hongqiu Wu\textsuperscript{\rm 1}, Yan Wang\textsuperscript{$^*$}, Xingyuan Liu\textsuperscript{\rm 1}, Hai Zhao\textsuperscript{\rm 1\thanks{Corresponding authors}}, Min Zhang\textsuperscript{\rm 2}\\
  \textsuperscript{\rm 1}Department of Computer Science and Engineering, Shanghai Jiao Tong University\\
  \textsuperscript{\rm 2}Harbin Institute of Technology, Shenzhen\\
  \texttt{wuhongqiu@sjtu.edu.cn,yanwang.branden@gmail.com}
}

\colmfinalcopy 
\begin{document}
\maketitle

\begin{figure*}
\centering
\includegraphics[width=0.82\textwidth]{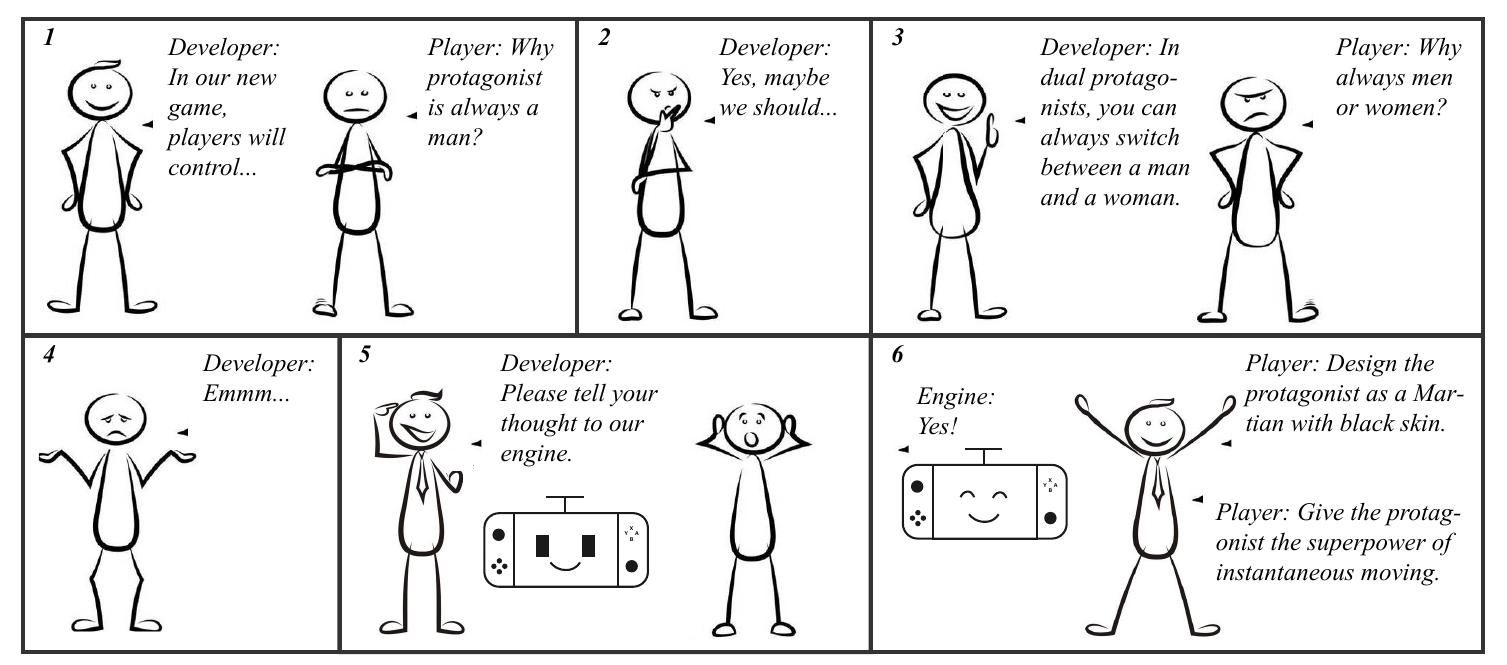}
\caption{1: Players were tired against the game's protagonist models. 2, 3: Developers thus created a new mode with dual protagonists. Players still didn't buy it, while they didn't know how to develop games. 4: There were irreconcilable divides between players and developers. 5, 6: Till the advent of the IDGE, it can read the players' mind and let them experience the games immediately.}
\label{f:comic}
\end{figure*}

\begin{abstract}
The \textit{Instruction-Driven Game Engine (IDGE)} project aims to democratize game development by enabling a large language model (LLM) to follow free-form game rules and autonomously generate game-play processes. The IDGE allows users to create games by issuing simple natural language instructions, which significantly lowers the barrier for game development. We approach the learning process for IDGEs as a \textit{Next State Prediction} task, wherein the model autoregressively predicts in-game states given player actions. It is a challenging task because the computation of in-game states must be precise; otherwise, slight errors could disrupt the game-play. To address this, we train the IDGE in a curriculum manner that progressively increases the model's exposure to complex scenarios.

Our initial progress lies in developing an IDGE for Poker, a universally cherished card game. The engine we've designed not only supports a wide range of poker variants but also allows for high customization of rules through natural language inputs. Furthermore, it also favors rapid prototyping of new games from minimal samples, proposing an innovative paradigm in game development that relies on minimal prompt and data engineering. This work lays the groundwork for future advancements in instruction-driven game creation, potentially transforming how games are designed and played.\footnote{Demo: \url{https://www.bilibili.com/video/BV1dA4m1w7xr/}; \url{https://youtu.be/jHTluHxJhqE}}\footnote{Repo: \url{https://github.com/gingasan/idge}}
\end{abstract}

\section{Introduction}
Game developers dedicate creativity to offer immersive experiences to game players. Players immerse themselves in games and offer valuable feedback to developers. This makes a symbiotic relationship between creators and customers.
However, as depicted in Figure~\ref{f:comic}, significant disconnections persist, due to diverse preferences of players across age, gender, and cultural backgrounds. Despite the fact that many today's games allow for customization of characters and appearances, it is an impossible task for developers to craft every aspect of the game to suit the need of every player.
Our study seeks to reconcile such a divide.

Game engines, as the heart of game development, are conventionally driven by complex programming languages. This technical barrier often deters enthusiasts from realizing their game development dreams. In response, we propose a novel concept: \emph{Instruction-Driven Game Engine (IDGE)}. This engine is designed to be instructable and scalable, enabling anyone to fashion a new game simply by providing natural language instructions. Distinct from recent advancements in video-based Game AI, such as CRADLE~\cite{tan2024towards} and SIMA~\cite{sima}, our focus in this paper is on the text-based prediction of game states, and leveraging Unity to render these text-described states to visually display.

IDGE is a neural engine, meaning it is built upon neural networks, specifically large language models (LLMs) \citep{DBLP:conf/nips/BrownMRSKDNSSAA20,DBLP:journals/corr/abs-2303-08774,DBLP:journals/corr/abs-2307-09288,DBLP:journals/corr/abs-2309-10305}. It is designed to follow a \textbf{game script} - a detailed instruction that portrays the game settings, rules, elements, etc. - and drive the progression of game-play as interacting with players. IDGEs frame the operation of engines as a \textit{Next State Prediction} task, autoregressively predicting the next in-game state based on the user-specified game script, previous in-game state, and current player action.

Training an IDGE faces the dual challenges of \textbf{stability} and \textbf{diversity}. The former seeks to provide a stable and precise game-play throughout lengthy contexts, while the latter seeks to follow diverse preferences within large player base. This necessitates that IDGE must adeptly navigate through a diverse set of game scripts while driving the games stably. Unfortunately, we empirically see a somewhat ironic twist: the model trained directly from existing game data seems to be neither stable nor diverse. Therefore, we employ a standard-to-diverse curriculum learning methodology to gradually introduce complexity into the training process. This strategy is designed to incrementally enhance the model's diversity while preserving its stability.

While it is still on journey from building an IDGE capable of producing AAA games, this paper provides our initial progress on \textbf{Poker}, a worldwide card game, e.g. \emph{Texas hold'em}, \emph{Badugi}.
We train the IDGE using data sourced from a poker simulator.
We show that the IDGE's understanding of nuanced semantics successfully fills voids left by the simulator program, e.g. suits, numbers, and game flow that never occurred in the training process.
We further show its great potential to generalize to brand new game scripts, e.g. new card combinations and battle strategies, by few-shot learning and continue learning, a bold and promising idea for future game development process.

We summarize our paper below:
$\bullet$ $\S$ \ref{s2} presents the concept of IDGE and formulates it as a learnable next state prediction task;
$\bullet$ $\S$ \ref{s3} discusses the game-style data for the poker engine;
$\bullet$ $\S$ \ref{s4} proposes the specific training techniques to enhance the training.

\section{Game Engine}
\label{s2}
In this section, we introduce the dialogue-style LLM as the setup for \emph{IDGEs}. Then, we discuss how we formulate the learning of IDGEs as a \textit{Next State Prediction} problem.

\subsection{From Instruction-Driven Dialogue to Instruction-Driven Game Engine}
LLMs shape their knowledge by navigating massive amounts of human data.
Most LLMs have been fine-tuned on dialogue-style corpora, where they are endowed with the ability to interact with users.
The resultant models, such as ChatGPT \citep{DBLP:journals/corr/abs-2303-08774}, can follow a system instruction provided by users, and generate responses in line with the instructions during interaction.

Likewise, a game engine works through interaction, too. For an IDGE, the system instruction specifically refers to a game script that accurately describes the desired game. In game-play, the IDGE follows the predefined game script and interacts with players, concurrently processing player inputs, such as moves and targets, to dynamically generate the in-game states as responses.

\begin{figure*}[t]
\centering
\includegraphics[width=0.86\textwidth]{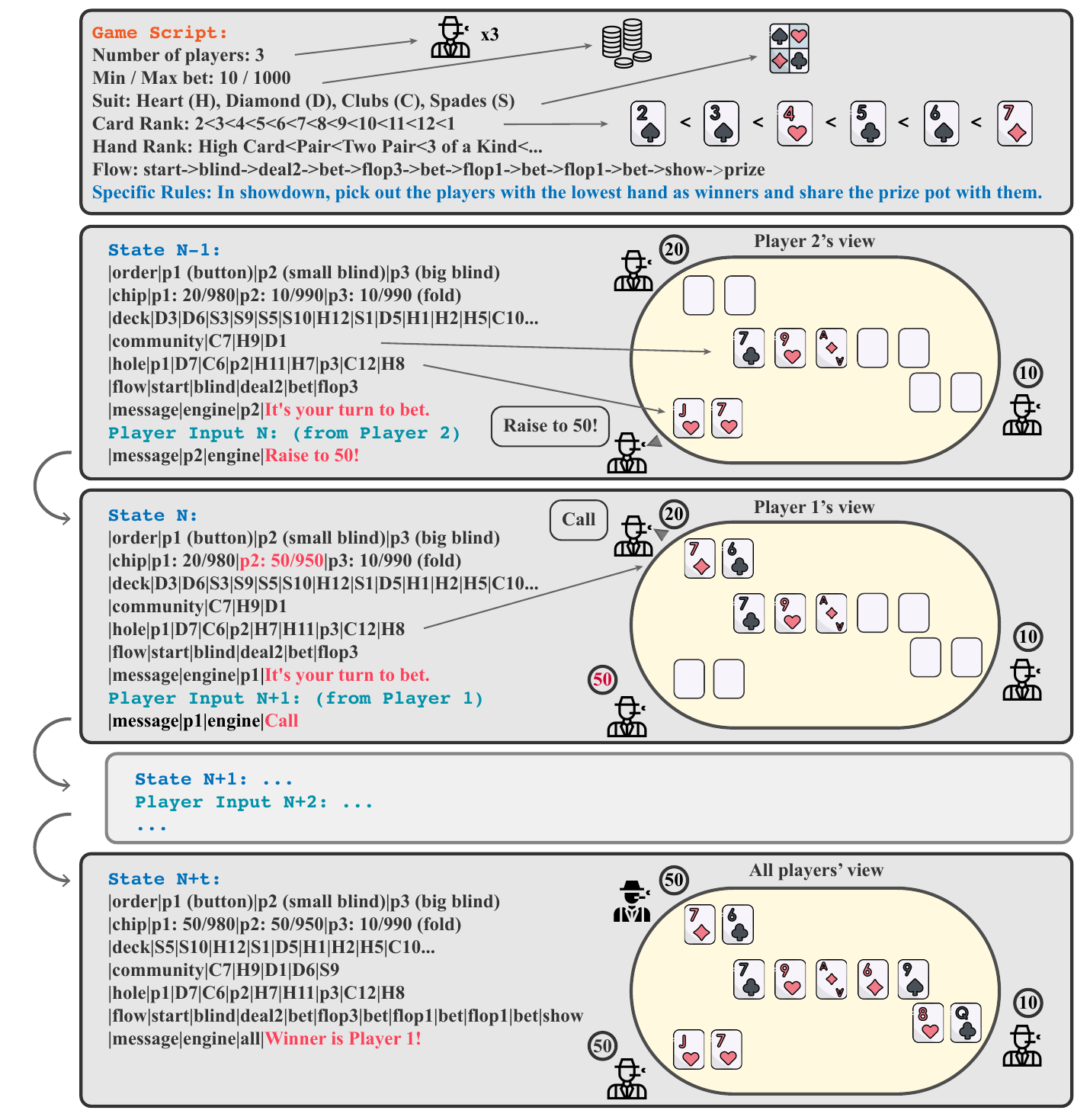}
\caption{Game-style samples for next state prediction. The left side is the input text for the engine from a global view, including all parts that are visible to players as well as those that are not. The right side is the diagram of the game from different views.}
\label{f:ex}
\end{figure*}

In Figure~\ref{f:ex}, we demonstrate how a poker IDGE facilitates a variant of \textit{Texas hold'em}: the user (or player) first inputs the game rules as the game script, with some specific rules described in natural language (described in the ``Specific Rules'' part),
with a variation from standard \textit{Texas hold'em} game.
In game-play, following this game script, the IDGE computes and returns the game-play process state by state, with player actions, e.g. check, call, raise, till the game concludes. More technical details about how we infer these states will be introduced in $\S$ \ref{s4}.

\subsection{Next State Prediction}

Causal language models learn the interplay of words and phrases through the autoregressive process of next token predicting \citep{DBLP:conf/nips/VaswaniSPUJGKP17,DBLP:conf/nips/BrownMRSKDNSSAA20}.
From a game-play perspective, the minimum component is no single token, but rather each in-game state.
An in-game state is a single frame that contains all current game status, e.g. characters, items, and missions. Essentially, the task of any game engines is exactly to compute the next state according to prior ones. Therefore, we may formulate the learning of IDGEs as a \emph{Next State Prediction (NSP)} problem.

Given a sequence of in-game states $\mathbf s=\{s_0,s_1,\cdots,s_T\}$, an IDGE with parameters $\theta$ seeks to maximize the likelihood:
$\sum_{t=1}^{T}\log p_\theta(s_t|s_0,s_1,\cdots,s_{t-1},x_t,z)$ 
where $x_t$ refers to the player input at time $t$, and $z$ refers to the game script which is global for the entire game.
The engine seeks to predict the next state given the prior states following $z$.

An in-game state is typically much bigger than a single token, incurring overflow of input and weakness of long-range capture for language models \citep{DBLP:journals/corr/abs-2004-05150,DBLP:journals/corr/abs-2309-17453}.
A relaxed case occurs when it is assumed that each state $s_t$ solely depends on its previous $k$ states.
Specifically when $k=1$, the former equation can be reduced to:
\begin{equation}
    \sum_{t=1}^{T}\log p_\theta(s_t|s_{t-1},x_t,z).
    \label{e2}
\end{equation}

While such an independence assumption would incur information loss, fortunately for a game engine, this loss can be avoided by the design of in-game states.
We will discuss a concrete example in the following section.

\section{Data for IDGE}
\label{s3}

In this paper, we generate a large number of game logs from a poker simulator. The simulator primarily supports ten representative types of poker games: \textit{Texas hold'em}, \textit{5-card draw}, \textit{Omaha}, \textit{Short-deck hold'em}, \textit{2-to-7 triple draw}, \textit{A-to-5 triple draw}, \textit{2-to-7 single draw}, \textit{Badugi}, \textit{Badeucey}, and \textit{Badacey}. Additionally, it allows for further configuration of several common elements for each poker game, including the number of players, the types of suits, the ranking of single-cards, the ranking of multi-card combinations, minimum and maximum bet limits, and the game flow, as shown in Figure \ref{f:ex}.
Detailed explanations of these functions can be found in Appendix \ref{a:func}. By adjusting these common elements, one can derive virtually infinite variations beyond the aforementioned ten representative poker games. Importantly, in our dataset, each game corresponds to a unique configuration, which augments the model's ability to follow various game scripts.

Moreover, we realize that if the game logs are sampled completely in uniform, the occurrence of some rare states, e.g. card combinations, would be extremely low. The resultant engine trained on such data may fall short in low-frequency situations, even though the dataset is large. Therefore, we balance the data by up/down-sampling the game logs to ensure that all possible situations occur similarly.
We show a concrete instance of our balancing process in Appendix \ref{a:bl}.

After obtaining game logs, we transform each log into a training sample as shown in Figure \ref{f:ex}, for Next State Prediction (NSP). Each sample is made up of three parts: the game script $z$ (structured or natural language), player input $x_t$, and in-game states $s_t$. If we were to draw an analogy with ChatGPT, these three parts respectively play the roles of the system, user, and assistant.

\noindent$\spadesuit$ \textbf{Game Script}\;
We design a structured template for the game script to represent the customized configuration of each game, as depicted by the black text in the top part of Figure~\ref{f:ex}.

In addition to the structured script, another part of the game script includes the description of specific game rules in natural language. The structural script can fully describe the variants of only two out of the ten types of games: \textit{Texas Hold'em} and \textit{5-card draw}. For the remaining eight types of games, we utilize natural language to describe the aspects not covered by the structural script. 
For instance, the blue text within the top part of Figure~\ref{f:ex} corresponds to the manually written specific rules of the game \textit{2-to-7 triple draw}. Additionally, for each game, we use GPT3.5 to paraphrase its description, to enhance the IDGE's eventual generalizability to different game scripts. The prompt we use is in Appendix \ref{a:prompt}.

\noindent$\spadesuit$ \textbf{In-Game State and Player Input}\;
For the in-game state and player input, we design a standardized language to represent them precisely and in short. As shown in the left side of Figure \ref{f:ex}, ``$\vert$deck'' is followed by the remaining cards in the pile, while ``$\vert$message$\vert a\vert b$'' is followed by the message sent from $a$ to $b$. Specifically, player 2 chooses to raise the bet. Given both the previous states and player action as input, the engine outputs a new state, where the chips of player 2 are updated and player 1 is informed to bet since player 3 has folded.

To ensure the independence assumption that each state $s_t$ solely depends on its previous state $s_{t-1}$, we design the in-game state to include all the information required for computing the next state.
As a result, regardless of the amount of history information, the engine can always precisely represent the game status by updating the above elements.

\begin{table}[]
\centering
\small
\caption{Statistics of training data.}
\label{t:dataset}
\begin{tabular}{rrrrrrrrrr}
\# of   & \# of       & \# of      & \# of   & avglen.   & avglen.    & vocab & avg.     & avg. bet    & avg.        \\
samples  & structured & natural    & poker   & (script)  & (state)    & size  & players  & (min/max)   & states      \\ \toprule
100k     & 49.8k      & 50.2k      & 10      & 208.4     & 306.3      & 1120  & 4.6      & 6.0/1841    & 35.3        \\
\end{tabular}
\end{table}

\noindent\textbf{Data Statistics}\;
Table \ref{t:dataset} shows the eventual statistics of the training data. We obtain 100k samples from the simulator, where structured and natural language scripts are nearly equal. 
The average number of states of one simulated round is 35.3, i.e. the number of states for the engine to predict.

\section{Curriculum Learning}
\label{s4}
Straightforwardly, we could utilize the standard data generated in $\S$ \ref{s3} to fine-tune a base model by maximizing Eq. \ref{e2} and obtain the IDGE. However, it may struggle with stability and diversity: neither can it accurately predict the next state nor comprehend the game script specified by users in natural language. Therefore, we devise a progressive curriculum learning process \citep{DBLP:conf/icml/BengioLCW09}, to incrementally enhance the IDGE's diversity while preserving stability.

\noindent\textbf{\textsc{Warmup}: Training with Core Set}\;
A game engine encompasses a complex collection of functionality. For example, a poker engine should deal, flop, and switch cards with deftness. The cold start problem is highlighted when all these sub-tasks are thrust into the model at once.
To this end, we propose a pre-learning phase to warmup the engine.
We define a \emph{Core Set (CS)}, a collection of fundamental functions that form the backbone capabilities of the engine, e.g. dealing a certain number of cards, listing all possible combinations from the given cards.
We derive an instruction tuning dataset from it, where each sample is a single instruction of a poker function.
We heuristically craft 40 functions in the core set (see Appendix \ref{a:core}).

The core set is akin to a basic library in most computer systems, while each function is an instruction described in natural language and the LLM learns these functions as a way of instruction following.

\noindent\textbf{\textsc{Standard}: Training on Standard Game Scripts}\;
The next step is to train the model on the standard data introduced in $\S$ \ref{s3} by optimizing NSP.
In this process, the task is to learn to become an engine by following the game scripts, combining pre-learned core functions organically.

\noindent\textbf{\textsc{Diverse}: Training on Rephrased Game Scripts}\;
While our standard data already includes some game scripts with natural language description, the majority of game scripts are still structured in nature. Mastering structured description can be cumbersome and too strict for users. Rather, it is more natural for them to describe the desired game in natural language. Rather than exhaustively crafting new natural language data, we propose \emph{Segment Rephrasing (SR)}, a technique that rephrases a portion of the structured description into natural language to encourage the model to follow diverse natural language game scripts.

As we introduced in $\S$ \ref{s3}, the structured part of the game script contains seven elements. We randomly sample several of them and rephrase them into natural language. An example is shown in Table~\ref{t:rephrase} in Appendix.
To largely keep the semantics intact, there is only a very low probability that the entire script will be rephrased.
The rephrasing process is done by GPT3.5 (in Appendix \ref{a:prompt}).
These rephrased game scripts will be more challenging to understand.
The resultant model acquires the capability to follow sophisticated game scripts accurately (natural language or a mixture of structured and natural language). Readers may refer to Table \ref{t:ood} in $\S$ \ref{s6} for a concrete example.

We eventually summarize the training pipeline as follows:
1) train on a 10k core set $\mathcal D_{E}$; 2) train by optimizing NSP on the standard dataset $\mathcal D_M$ (100k); 3) train by optimizing NSP on 10k segment rephrased samples $\mathcal D_H$ and 10k standard samples.




The \textbf{warmup}, \textbf{standard}, and \textbf{diverse} process correspond to the easy, medium, and hard curriculum.
It is a smooth transfer of the IDGE from standardization to fully instructable.

\section{Experimental Results}
\label{s6}

We evaluate the IDGE on two scenarios: in-domain and out-of-domain games. The former is automatically generated by our simulator, which can be considered as a test set that has the same distribution as training. The latter resembles real-world situations, where proficient poker players are directly enlisted as annotators to create new game scripts. Subsequently, games are played based on these new scripts to obtain test data.
Ablation study is reported in Appendix \ref{a:abl}.

\subsection{Training and Evaluation Setup}
We train each model using LoRA \citep{DBLP:conf/iclr/HuSWALWWC22} with $r=8$, $\alpha=32$, and the optimal learning rate in 1.5e-4 and 3e-4.
We set the warmup steps to 30 and total batch size to 8 for 8 chips.
For each curriculum, we train 5 epochs.
To ensure the stability of outputs, we leverage greedy decoding.

\begin{figure*}
\centering
\includegraphics[width=0.85\textwidth]{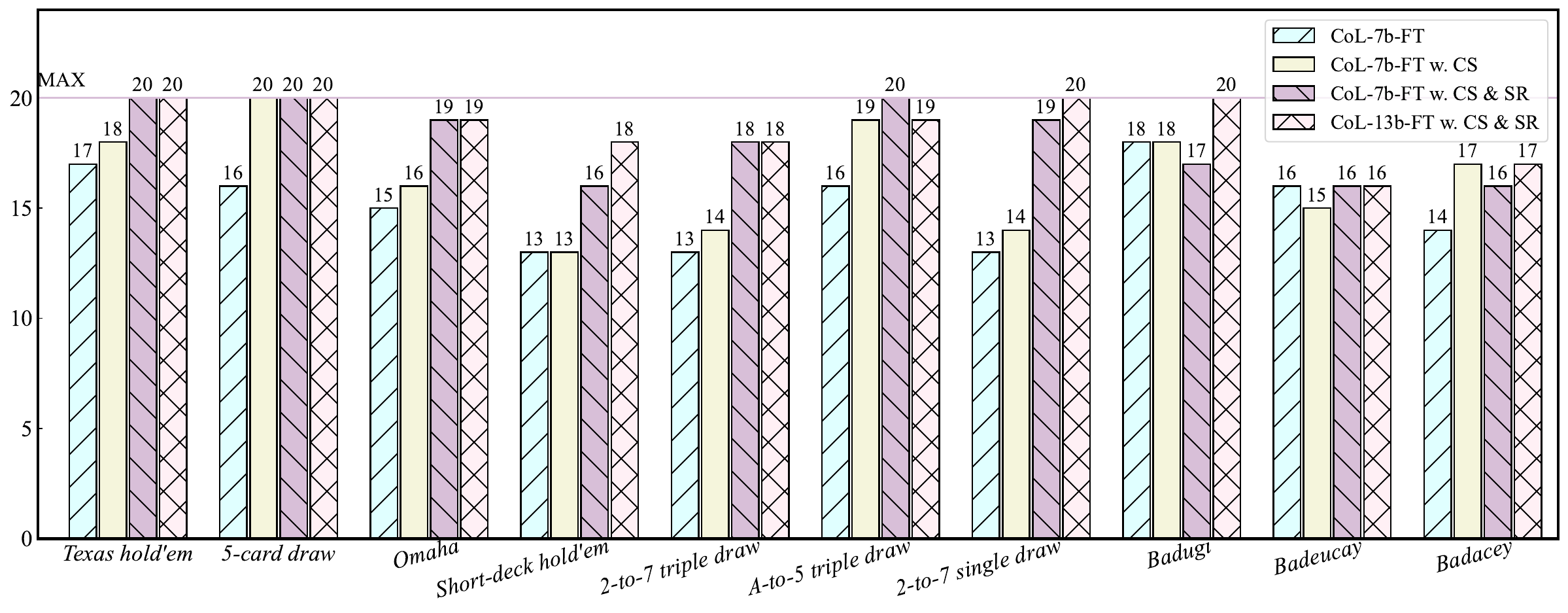}
\caption{Round-level success rates on variants of 10 prototype poker games. In addition, our experiment turns out both GPT3.5 and GPT4 only achieve a zero success rate on all these games.}
\label{f:id-round}
\end{figure*}

$\bullet$ \textbf{In-domain evaluation}: The model has been exposed to a broad range of variants based on ten existing poker games during training. We sampled some unseen variants of these ten games from the poker simulator for evaluation. Then, we program some random players that randomly select an action as their input, to play games on IDGE. In such a scenario we can quickly evaluate the correctness of IDGE's predicted states using the poker simulator. There are 200 games in total in our in-domain evaluation set.

$\bullet$ \textbf{Out-of-domain evaluation}: The in-domain game scripts allow for the configuration of only seven key elements. To evaluate our engine performance in scenarios more closely aligned with the real world, we further recruit five proficient poker players as our engine testers. Each of them is asked to create a new game based on their personal preferences using natural language. They are free to tailor the game scripts, but these scripts should not be entirely unrelated to \emph{Texas hold'em} or \emph{5-card draw}. Subsequently, we invite them to play ten rounds for each new game and record all player inputs and in-game states throughout the game-play. This forms our out-of-domain evaluation set that comprises five distinct game scripts and 50 rounds of games.

\subsection{In-Domain Games}
\noindent\textbf{Round-level}\;
Figure \ref{f:id-round} depicts the round-level success rates of fine-tuned CodeLLaMA \citep{DBLP:journals/corr/abs-2308-12950} (CoL).
The success rate is counted if the engine correctly handles all states in a round.

In our results, the most surprising aspect is the performance of GPT3.5 and GPT4. In 200 rounds of games, they are unable to successfully complete any single round. In contrast, fine-tuned models produce satisfactory results. Even the vanilla CoL-7b-FT model achieves an average success rate. Utilizing the core set (CS) to warmup further improves the performance, outperforming the vanilla one in seven out of ten game variants. Furthermore, the model underwent full curriculum (w. CS \& SR) showcases a remarkable improvement, e.g. +5 on \emph{2-to-7 triple draw} and +6 on \emph{2-to-7 single draw}. This suggests that challenging rephrasing samples encourage the model to learn a better alignment between structured and natural language game scripts, thereby better balancing stability and diversity. Lastly, increasing the model size also leads to performance gain.
\begin{table*}[t]
\caption{State-level performances with different training methods. $\star$ refers to the model w. CS \& SR.}
\label{t:id-func}
\centering
\small
\begin{tabular}{lrrrrrrrrr}
\toprule
                                     & \texttt{start} & \texttt{blind}& \texttt{deal} & \texttt{flop}  & \texttt{switch} & \texttt{bet}  & \texttt{show} & \texttt{prize} & avg.     \\ \midrule
GPT3.5 - 5-shot                      & 74.0           & 60.5          & 0             & 6.0            & 6.9             & 46.0            & 40.5        & 78.5           & 39.1     \\
GPT4 - 5-shot                        & 68.0           & 84.0          & 0             & 0              & 11.7            & 20.6            & 48.0        & \textbf{92.0}  & 40.5     \\
CodeLLaMA-7b - FT
                                     & 100            & 100           & 90.0          & 100            & 98.9            & 98.3          & 80.5          & 99.0           & 95.8     \\
\;\;w. \emph{Core Set Warmup}        & 100            & 100           & \textbf{100}  & 100            & 99.6            & \textbf{99.7} & 91.0          & 100            & 98.8     \\
\;\;\;w. \emph{Segment Rephrasing}   & 100            & 100           & 100           & 100            & \textbf{100}    & 98.5          & \textbf{92.0} & 100            & 98.8     \\
LLaMA2-7b - FT$^\star$
                                     & 100            & 100           & 98.0          & 100            & 96.0            & 96.6          & 78.5          & 96.0           & 95.6     \\
CodeLLaMA-13b - FT$^\star$           & 100            & 100           & 100           & 100            & 100             & 100           & 95.5          & 100            & \textbf{99.4} \\ \bottomrule
\end{tabular}
\end{table*}

\noindent\textbf{State-level}\;
One might question why GPT3.5 and GPT4 completely fail in this task, significantly behind fine-tuned CodeLLaMA. To conduct a more in-depth analysis, we compute the state-level accuracy corresponding to each function in Table~\ref{t:id-func}\footnote{\url{https://huggingface.co/meta-llama/Llama-2-7b-chat-hf};\url{https://huggingface.co/codellama/CodeLlama-7b-Instruct-hf};\url{https://huggingface.co/codellama/CodeLlama-13b-Instruct-hf}}.
We find that, though GPT3.5/4 are strong in mathematical calculation (\texttt{prize}), they perform poorly in remaining functions. The underlying fact is that they struggle to manage cards accurately.
For example, they are very likely to mess up the order of cards, hallucinating new cards or missing some of them.
We conjecture that they were not exposed to highly sophisticated data and tasks as for the engine during their training.
Accumulation of errors in all these functions leads non-fine-tuned models to zero success rates in round-level evaluation.
It is important to note that the engine is required to be all-round at each function; otherwise, the overall engine performance will degenerate in a way of \textbf{Buckets effect}.

For fine-tuned models, they perform close to 100\% in most functions, which guarantees the engine's stable round-level performance as in Figure \ref{f:id-round}.
CS shows particularly beneficial for two specific functions, i.e. +10 on \texttt{deal}, +10.5 on \texttt{show}, which we notice as challenging aspects for a poker engine.
We also find that code pre-training is beneficial to IDGEs, i.e. CodeLLaMA works better than LLaMA2.
We conjecture that processing structured inputs as in-game states shares similarity with processing code, which is also structured \citep{DBLP:conf/iclr/GuoRLFT0ZDSFTDC21,DBLP:conf/acl/WuZZ21}.

\subsection{Out-of-Domain Games}

\begin{table}[t]
\caption{Out-of-domain game scripts. We skip some settings for brevity, e.g. the number of players. Particularly for script 3, we did not include the ``All-in'' operation in the training data.}
\label{t:ood}
\centering
\small
\begin{tabular}{l@{}}
\toprule
\textbf{Script 1: \emph{reverse ranking + less suits}} \\ \cmidrule[1.2pt](l{0pt}r{200pt}){1-1}
\begin{tabular}[c]{@{}l@{}}In this game of Texas hold'em, there are 3 players and the minimum bet is 2, maximum bet is 1000.\\ \textbf{The card numbers rank} from low to high as follows: 1, 13, 12, 11, 10, 9, 8, 7, 6, 5, 4, 3, 2.\\ \textbf{The suits are D, O, and G.}\end{tabular} \\ \midrule[0pt]
\textbf{Script 2: \emph{additional dealing phase}} \\ \cmidrule[1.2pt](l{0pt}r{200pt}){1-1}
\begin{tabular}[c]{@{}l@{}}The game proceeds in the following order: start, blinds, deal 2 cards, bet, reveal 3 cards (the flop), bet,\\ reveal 1 card (the turn), \textbf{deal 1 cards (new deal)}, bet, show, and finally the prize is distributed.
\end{tabular} \\ \midrule[0pt]
\textbf{Script 3: \emph{all-in}} \\ \cmidrule[1.2pt](l{0pt}r{200pt}){1-1}
\begin{tabular}[c]{@{}l@{}}Define a new player operation in the phase of bet ``All-in''. The player puts all his remaining chips into\\ the pot, and is no longer able to make further bet during the game.
\end{tabular} \\ \midrule[0pt]
\textbf{Script 4: \emph{3-card draw}} \\ \cmidrule[1.2pt](l{0pt}r{200pt}){1-1}
\begin{tabular}[c]{@{}l@{}}Introduce a new game, named ``3-card draw''. In this game, there are 3 suits, H, D, C.\\
In addition, define \textbf{two new combinations with 3 cards in hand.}\\
``Straight'': there are three consecutive cards, e.g. C10, H11, D12.\\
``Flush'': there are three cards within the same suit, e.g. H1, H10, H6.\\
All combinations rank as:
High Card$<$Pair$<$Three of a Kind$<$Straight$<$Flush$<$Straight Flush. \end{tabular} \\ \midrule[0pt]
\textbf{Script 5: \emph{6-card draw}} \\ \cmidrule[1.2pt](l{0pt}r{200pt}){1-1}
\begin{tabular}[c]{@{}l@{}}Introduce a new game, named ``6-card draw''. In this game, there are 5 suits, H, D, C, S, and R.\\
In addition, define \textbf{two new combinations with 6 cards in hand.}\\
``\textbf{Three Pair}'': there are three pairs of distinct numbers, e.g. R8, H8, C10, H10, H12, D12.\\
``\textbf{Big House}'': there are two pairs of three of one kind, e.g. R8, H8, C8, D12, H12, D12.\\
All combinations rank as:\\
High Card$<$Pair$<$Three of a Kind$<$Straight$<$Flush$<$\textbf{Full House$<$Three Pair$<$Big House}$<$Straight Flush \end{tabular} \\
\bottomrule
\end{tabular}
\end{table}

\begin{table}[]
\centering
\small
\caption{Success rates for 10 rounds (\%) on out-of-domain games. CT refers to continue-training.}
\label{t:ood-round}
\begin{tabular}{lrrrrrrrr}
\toprule
                            & \multicolumn{4}{l}{w/o. \emph{Segment Rephrasing}} & \multicolumn{4}{l}{w. \emph{Segment Rephrasing}} \\
\textbf{Script}                         & \textbf{0-shot} & \textbf{3-shot} & \textbf{10-shot}    & \textbf{CT}   & \textbf{0-shot}    & \textbf{3-shot} & \textbf{10-shot} & \textbf{CT}            \\ \midrule
\textbf{1:} \textit{ranking + suits}    & 80                 & 100             & -                & -             & \textbf{100}       & 100             & -                & -                      \\
\textbf{2:} \textit{additional dealing} & 0                  & 70              & -                & -             & \textbf{80}        & 100             & -                & -                      \\
\textbf{3:} \textit{all-in}             & -                  & 0               & 10               & -             & -                  & \textbf{70}     & 100              & -                      \\
\textbf{4:} \textit{3-card draw}        & -                  & 0               & 0                & -             & -                  & 50              & \textbf{80}      & \textbf{100} (8-shot)  \\
\textbf{5:} \textit{6-card draw}        & -                  & 0               & 0                & 0             & -                  & 0               & 10               & \textbf{100} (23-shot) \\ \bottomrule
\end{tabular}
\end{table}

In Table~\ref{t:ood}, from script 1 to 5, customization of the script becomes more, and the gap from standard scripts also becomes larger.
For example, the most challenging script 5 portrays a brand new game with defining two novel six-hand combinations, ``Three Pair'' and ``Big House''.

We apply few-shot in-context learning to adapt the engine to new games. The in-context samples are also crafted by invited players.
Instead of listing all samples of a complete round, which leads to lengthy input, we allocate them by their functions. In other words, for each function, we solely place its corresponding samples in its game script (example in Appendix \ref{a:icl}).

We report the results in Table \ref{t:ood-round} based on CoL-13b with and without SR.
Intuitively, models incorporated SR significantly outperform those without SR across all five game scripts, suggesting that SR is indispensable for the engine to understand pure natural language inputs.
Script 3 and 4 present two challenging cases, as they define new operations or combinations. We observe that even models enhanced with SR struggle to accurately compute each state solely based on the scripts. Fortunately, when we provide it with some samples, the engine operating in a few-shot learning manner achieves quite satisfactory results. This is exactly a novel game development process introduced by IDGEs: we may shift the burden from writing new programs to crafting a detailed game script and a small number of samples, a process often referred as ``prompt engineering''.

Script 5 presents the most challenging case, pushing the engine's upper limit. Due to its complexity, the model can no longer accurately predict solely through in-context learning. Therefore, we explore a new solution: \textbf{continue-training}.
We ask players to manually label incorrectly predicted states while playing script 5 using the IDGE, and correct them before continuing the game. In this manner, we construct a number of pairs of ``good states'' and ``bad states''. We thus employ DPO~\citep{rafailov2023direct} to optimize the engine. We find that it quickly learns to rectify its shortcomings guided by user feedback, eventually achieving satisfactory outcomes (achieving 100\% with 8/23-shot for script 4/5). This provides a new perspective for application of IDGEs in real-world scenarios: users can customize their individual evolving IDGE by own feedback, continually refining it till satisfied.

\section{Conclusion}
This paper introduces the Instruction-Driven Game Engine, offering game enthusiasts a brand new game development and playing experience. We formulate the learning of IDGEs as Next State Prediction and leverage a curriculum learning approach to enhance stability and diversity. Experiments demonstrate the proposed engine can accurately complete the majority of user-defined games. For challenging cases, the engine can evolve with user feedback in an RLHF manner.

This paper presents our initial progress in Poker games. Such a paradigm theoretically applies to all types of games. However, our progress is constrained by several bottlenecks:

\textbf{Inference Latency} We have demonstrated that IDGEs go well with turn-based strategy (TBS) games. For real-time strategy (RTS) games, players may make more than one action per second. The inference latency of LLMs cannot meet the real-time requirements of such games.

\textbf{Context Window Size} Generally, as games become more complex, the length of in-game states increases, posing a challenge to satisfy our independent assumption. This may significantly challenge both the comprehension ability of LLMs and the cache of KV states.

\textbf{Accessibility} The kernel data of most commercial games is not publicly available, which is why we developed a poker simulator to generate the training data for this paper.

We are delighted to observe that there have been continuous advancements in inference frameworks such as vLLM~\citep{kwon2023efficient}, as well as efficient long-text generation methods like StreamingLLM~\citep{DBLP:journals/corr/abs-2309-17453} and Temp-LoRA~\citep{wang2024greater}. We believe that the ongoing development of LLM technologies will ultimately address the limitations of latency and the context window. Regarding the issue of accessibility, we look forward to more companies providing open interfaces as SC2LE~\citep{vinyals2017starcraft}, HOK Arena~\citep{wei2022honor} to offer kernel game data.



\bibliography{colm2024_conference,custom}
\bibliographystyle{colm2024_conference}

\appendix
\newpage
\section{Segment Rephrasing}

\begin{table}[]
\caption{An example of segment rephrasing.}
\label{t:rephrase}
\centering
\small
\textbf{Game Script for \emph{Texas hold'em}}\\
\begin{tabular}[c]{l@{}}
\midrule
Number of players: 3\\
({\color[HTML]{cc0066}In this game of Texas hold'em, there are three players.})\\
Card rank: 2$<$3$<$4$<$5$<$6$<$7$<$8$<$9$<$10$<$11$<$12$<$1\\
Min / Max bet: 10 / 5000\\
Flow: start-$>$blind-$>$deal2-$>$bet-$>$flop3-$>$bet-$>$flop1-$>$...\\
({\color[HTML]{cc0066}The game begins with placing the blinds, followed by}\\
{\color[HTML]{cc0066}dealing 2 cards to each player, placing the bet for each...})\\
\bottomrule
\end{tabular}
\end{table}

\section{Related Work}
A game engine is a fundamental software designed for the creation of games, providing developers with necessary tools.
Famous game engines include Unreal, Unity, CryENGINE, etc.
We spotlight two crucial properties of a game engine.
The first is functionality, i.e. providing a wide variety of basic tools to facilitate the development process.
The next is secondary development, i.e. rich and flexible interfaces to allow developers to customize new games.
In this work, we introduce a new concept, instruction-driven game engine (IDGE), a neural game engine learned on basis of large language models \citep{DBLP:journals/corr/abs-2303-08774,DBLP:journals/corr/abs-2307-09288,DBLP:journals/corr/abs-2310-06825,DBLP:journals/corr/abs-2309-10305,DBLP:journals/corr/abs-2304-08354}.
As opposed to a typical game engine, an IDGE acquires its functionality power by instruction tuning on the core set \citep{DBLP:conf/nips/BrownMRSKDNSSAA20,DBLP:journals/jmlr/RaffelSRLNMZLL20,DBLP:conf/nips/Ouyang0JAWMZASR22} and allows for low-barrier game development by natural language description.

AI for games is an exciting area in AI research.
A great amount of recent work studies learning for agents, e.g. as game players for Atari \citep{DBLP:journals/corr/MnihKSGAWR13}, Minecraft \citep{DBLP:conf/nips/FanWJMYZTHZA22,DBLP:journals/corr/abs-2305-16291}, StarCraft, \citep{DBLP:journals/nature/VinyalsBCMDCCPE19}, NetHack \citep{DBLP:conf/nips/KuttlerNMRSGR20,DBLP:conf/iclr/Lowe0FKP20}, Werewolf \citep{DBLP:journals/corr/abs-2309-04658}; as non-play characters (NPCs) \citep{DBLP:journals/nature/ShanahanMR23,DBLP:journals/air/UludagliO23}; player assistants \citep{gallotta2024large}; game commentators \citep{DBLP:conf/icids/Eladhari18,DBLP:conf/exag/RanellaE23}.
Most of these work aims to train an AI player based on game logs or images. However, our work diverges from all of them in that we focus on the LLM as a game engine, attempting to build a game engine that is defined by instructions (game scripts) and data (in-game states).
In addition, an agent focuses on the way AI behaves, while an engine focuses on the way AI would react in the face of any possible behaviors from any human being or agent.
More recent work comes up with learning for a foundation agent, a single agent with generalizable skills to behave in various environments,
e.g. SIMA \citep{sima}, an instruction-driven agent proficient in multiple simulated environments; CRADLE \citep{tan2024towards}, a powerful agent capable of playing complex AAA games like Red Dead Redemption 2 by controlling the keyboard and mouse.
However, our work targets the IDGE for a specific group of games, Poker, as an initial step for building a foundation IDGE.
Poker is a widely studied information game of immense popularity \citep{DBLP:journals/cacm/BowlingBJT17,DBLP:journals/corr/MoravcikSBLMBDW17,DBLP:journals/corr/abs-2308-12466,DBLP:journals/corr/abs-2308-07327,DBLP:conf/aaai/ZhaoYLLX22}.

In this paper, the entire training cycle for IDGE is a way of curriculum learning \citep{DBLP:conf/icml/BengioLCW09}. Recent studies show the promising role of curriculum learning in empowering the language models to tackle more challenging tasks \citep{DBLP:conf/emnlp/VakilA23,DBLP:conf/emnlp/Wu0ZZ23}.
The proposed segment rephrasing technique is related to perturbation training \citep{DBLP:conf/iclr/ZhuCGSGL20,DBLP:conf/iclr/WuLSZZ23,DBLP:journals/corr/abs-2310-05914}, which smooths the boundary of structured and natural language in the semantic space.

\section{Prompt used for GPT3.5}
\label{a:prompt}

\begin{table}[h]
\centering
\small
\caption{Prompts for rephrasing structural and natural language.}
\label{t:prompt}
\begin{tabular}{l}
\toprule
\begin{tabular}[c]{@{}l@{}}\textbf{\texttt{Prompt for rephrasing structural language}}\\ You are a skilled English writer.\\ I will give you a paragraph that describes the rules of a Texas hold'em game, split by triple backticks.\\ \\ \textasciigrave\textasciigrave\textasciigrave\\ \{sent\}\\ \textasciigrave\textasciigrave\textasciigrave\\ \\ 1. Use natural language to describe the rule.\\ 2. It is important that do not change the cards.\\ 3. You must reflect the ranking of the cards and do not change their order. This is also important. \\ 4. Do not include additional information that is not reflected in the sentence.\\ 5. Return the new sentence split by triple backticks.\\ 6. Do not use ``ace'' in your response.\\
\textbf{\texttt{Input example 1}}\\
Flow: start-$>$shuffle-$>$blind-$>$deal5-$>$bet-$>$switch-$>$bet-$>$switch-$>$bet-$>$show-$>$prize\\
\textbf{\texttt{Output example 1}}\\
Process: The game starts with shuffling the cards, followed by assigning blinds, dealing five initial\\ cards, placing bets, switching cards, placing more bets, switching again, placing bets once more,\\ revealing the cards, and finally awarding the prize.\\
\textbf{\texttt{Input example 2}}\\
Card Rank: 7$<$6$<$2$<$14$<$9$<$11$<$4$<$17$<$16$<$1$<$8$<$3\\
\textbf{\texttt{Output example 2}}\\
The card numbers, from lowest to highest value, are 7, 6, 2, 14, 9, 11, 4, 17, 16, 1, 8, 3, and 10.\end{tabular} \\ \midrule
\begin{tabular}[c]{@{}l@{}}\textbf{\texttt{Prompt for rephrasing natural language}}\\ Paraphrase the given paragraph.\\ \\ \textasciigrave\textasciigrave\textasciigrave\\ \{sent\}\\ \textasciigrave\textasciigrave\textasciigrave\\ \\ 1. It is important to keep your response as diverse as the input.\\ 2. You should change the structure of the paragraph in your response.\\
\textbf{\texttt{Input example 1}}\\
In showdown, pick out the players with the lowest combination of cards as the winners.\\
\textbf{\texttt{Output example 1}}\\
In showdown, winners are determined as those players with the lowest aggregates of hand.\\
\textbf{\texttt{Input example 2}}\\
In showdown, define a new ranking strategy called \textasciigrave Badugi \textasciigrave. In given cards, pick out the cards of\\ distinct suits and no pair. If there are more than one cards of the same suit or same value, choose the\\ smaller-ranking one. Hence, the greatest Badugi refers to the the most number of distinct cards and\\ the smallest values.\\
\textbf{\texttt{Output example 2}}\\
In the heat of showdown, draft a ranking outline known as \textasciigrave Badugi \textasciigrave. Choose cards that differ in\\ suit and lack matching values. When selecting between cards of identical suits or values, opt for\\ those with a lower rank. Hence, the epitome of a Badugi hand holds the most varied suits at the\\ smallest card values.\end{tabular} \\ \bottomrule
\end{tabular}
\end{table}

\section{Data Efficiency}
\label{a:bl}

\begin{figure}[h]
\centering
\includegraphics[width=0.69\textwidth]{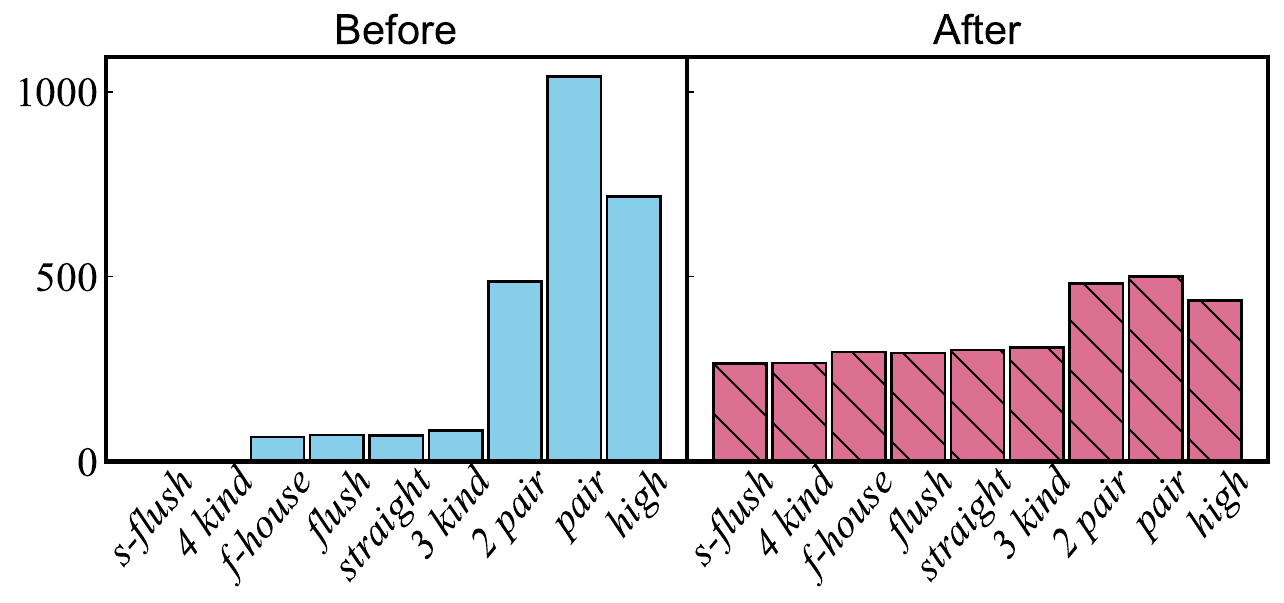}
\caption{Balance the data distribution to boost data efficiency when sampling training data.}
\label{f:manipulate}
\end{figure}

While the simulator program can generate massive data in a short period, we stress the importance of data efficiency.
We find that a simple manipulation on the real data distribution effectively allows for a better training outcome with a smaller amount of training data.
We use the combination of cards as an instance.
Typically in a poker game, the chance of a straight (e.g. 5, 6, 7, 8, 9) in hand is much lower than a pair (e.g. 6, 6).
As depicted in Figure \ref{f:manipulate} (left), recorded from 1,000 poker games, we find that the occurrences of ``pair'' and ``high card'' far outstrip the others.
The model trained on such unbalanced data may fall short in low-frequency situations, even though the dataset is large.
We thus balance the data sampling process so that each combination of cards occurs similarly in the game-play data, as in Figure \ref{f:manipulate} (right).

\section{Ablation Study}
\label{a:abl}

\begin{table*}[h]
\caption{Ablation study based on CoL-7b.}
\label{t:abl}
\centering
\small
\begin{tabular}{lrrrrrr}
& \texttt{start} & \texttt{deal} & \texttt{show} & \texttt{prize} & \textit{Script 1} & \textit{Script 2}     \\ \toprule
\begin{tabular}[c]{@{}l@{}}Warmup + Standard\\ \qquad w/o. data efficiency\end{tabular}
& 99.5           & 84.5          & 72.5          & 98.5           & 60.0              & 0             \\
Warmup + Standard
& 100            & 98.0          & 89.0          & 98.5           & 80.0              & 0             \\
\begin{tabular}[c]{@{}l@{}}Warmup + Standard + Diverse\\ \qquad w/o. down-sample\end{tabular}
& 100            & 89.0          & 86.5          & 100            & 90.0              & \textbf{90.0} \\
Warmup + Standard + Diverse
& 100            & \textbf{100}  & \textbf{92.0} & 100            & \textbf{100}      & 80.0          \\
\end{tabular}
\end{table*}

\section{Individual Function}
\label{a:func}
\definecolor{OxfordBlue}{rgb}{0.215,0.254,0.317}
\begin{longtblr}[
  caption = {Reference for each individual function.},
]{
  width = \linewidth,
  colspec = {Q[77]Q[860]},
  hline{1,11} = {-}{0.08em},
  hline{2} = {-}{},
}
\small
\textbf{Flow}          & \textbf{Meaning}                                                             \\
\texttt{start}         & Order the cards and players, including the button, and deliver their chips if needed.  \\
\texttt{blind}         & Place the blind bet, e.g. for the big blind and small blind.                 \\
\texttt{dealx}         & Deal x cards to each player.                                                 \\
\texttt{bet}           & Ask each player to place the bet into the prize pool one by one.             \\
\texttt{flopx}         & Skip one card before turning over x cards from the deck.                     \\
\texttt{switch}        & The player chooses to exchange some cards in his or her hand.                \\
\texttt{show}          & Show all hands of players and determine the winners.                         \\
\texttt{prize}         & Split the prize pool to the winners.                         
\end{longtblr}

\section{Game Script}
\label{script}
\begin{longtblr}[
  caption = {Game scripts of different poker games. We omit the number of players and bet limits.},
  label = {t:script},
]{
  width = \linewidth,
  colspec = {Q[944]},
  hlines,
  hline{1,21} = {1}{0.08em},
  hline{2-12} = {1}{0.05em},
}
\textit{\textbf{Texas~hold'em}}\\
{\textbf{Suit:} Hearts (H), Diamonds (D), Clubs (C), Spades (S)\\
\textbf{Card Rank:} 2\textless{}3\textless{}4\textless{}5\textless{}6\textless{}7\textless{}8\textless{}9\textless{}10\textless{}11\textless{}12\textless{}13\textless{}1\\\textbf{Hand Rank:}High Card\textless{}Pair\textless{}Two Pair\textless{}Three of a Kind\textless{}Straight\textless{}Flush\textless{}Four of a Kind\textless{}Straight Flush\\
\textbf{Flow:} start-\textgreater{}blind-\textgreater{}deal2-\textgreater{}bet-\textgreater{}flop3-\textgreater{}bet-\textgreater{}flop1-\textgreater{}bet-\textgreater{}show-\textgreater{}prize}\\
\textit{\textbf{5-card~draw}}\\
{\textbf{Suit:} Hearts (H), Diamonds (D), Clubs (C), Spades (S)\\
\textbf{Card Rank:} 2\textless{}3\textless{}4\textless{}5\textless{}6\textless{}7\textless{}8\textless{}9\textless{}10\textless{}11\textless{}12\textless{}13\textless{}1\\\textbf{Hand Rank:}High Card\textless{}Pair\textless{}Two Pair\textless{}Three of a Kind\textless{}Straight\textless{}Flush\textless{}Four of a Kind\textless{}Straight Flush\\
\textbf{Flow:}start-\textgreater{}blind-\textgreater{}deal5-\textgreater{}bet-\textgreater{}switch-\textgreater{}bet-\textgreater{}show-\textgreater{}prize}\\
\textit{\textbf{Short-deck~hold'em}}\\
{\textbf{Suit:} Hearts (H), Diamonds (D), Clubs (C), Spades (S)\\
\textbf{Card Rank:} 6\textless{}7\textless{}8\textless{}9\textless{}10\textless{}11\textless{}12\textless{}13\textless{}1\\
\textbf{Hand Rank:} High Card\textless{}Pair\textless{}Two Pair\textless{}Three of a Kind\textless{}Flush\textless{}Straight\textless{}Four of a Kind\textless{}Straight Flush\\
\textbf{Flow:} start-\textgreater{}shuffle-\textgreater{}deal2-\textgreater{}bet-\textgreater{}flop3-\textgreater{}bet-\textgreater{}flop1-\textgreater{}bet-\textgreater{}flop1-\textgreater{}bet-\textgreater{}show-\textgreater{}prize\\
\textbf{Specific Rules:}\\
Define a new combination ``Small Straight''. It allows the highest-ranking card to be used as the lowest-ranking when forming the straight. A small straight is smaller than a standard}\\
\textit{\textbf{Omaha}}\\
{\textbf{Suit:} Hearts (H), Diamonds (D), Clubs (C), Spades (S)\\
\textbf{Card Rank:} 2\textless{}3\textless{}4\textless{}5\textless{}6\textless{}7\textless{}8\textless{}9\textless{}10\textless{}11\textless{}12\textless{}13\textless{}1\\\textbf{Hand Rank:}High Card\textless{}Pair\textless{}Two Pair\textless{}Three of a Kind\textless{}Flush\textless{}Straight\textless{}Four of a Kind\textless{}Straight Flush\\
\textbf{Flow:} start-\textgreater{}blind-\textgreater{}deal4-\textgreater{}bet-\textgreater{}flop3-\textgreater{}bet-\textgreater{}flop1-\textgreater{}bet-\textgreater{}show-\textgreater{}prize\\
\textbf{Specific Rules:}\\
In showdown, only a combination of 2 hole cards and 3 community cards can be used to form the optimal cards.}\\
\textit{\textbf{2-to-7~triple~draw}}\\
{\textbf{Suit:} Hearts (H), Diamonds (D), Clubs (C), Spades (S)\\
\textbf{Card Rank:} 2\textless{}3\textless{}4\textless{}5\textless{}6\textless{}7\textless{}8\textless{}9\textless{}10\textless{}11\textless{}12\textless{}13\textless{}1\\
\textbf{Hand Rank:} High Card\textless{}Pair\textless{}Two Pair\textless{}Three of a Kind\textless{}Straight\textless{}Flush\textless{}Four of a Kind\textless{}Straight Flush\\
\textbf{Flow:} start-\textgreater{}blind-\textgreater{}deal5-\textgreater{}bet-\textgreater{}switch-\textgreater{}bet-\textgreater{}switch-\textgreater{}bet-\textgreater{}switch-\textgreater{}bet-\textgreater{}show-\textgreater{}prize\\
\textbf{Specific Rules:}\\
In showdown, pick out the players with the lowest combination of cards as the winners.}\\
\textit{\textbf{A-to-5~triple~draw}}\\
{\textbf{Suit:} Hearts (H), Diamonds (D), Clubs (C), Spades (S)\\
\textbf{Card Rank:} 1\textless{}2\textless{}3\textless{}4\textless{}5\textless{}6\textless{}7\textless{}8\textless{}9\textless{}10\textless{}11\textless{}12\textless{}13~~\\
\textbf{Hand Rank:} High Card\textless{}Pair\textless{}Two Pair\textless{}Three of a Kind\textless{}Full House\textless{}Four of a Kind\\
\textbf{Flow:} start-\textgreater{}blind-\textgreater{}deal5-\textgreater{}bet-\textgreater{}switch-\textgreater{}bet-\textgreater{}switch-\textgreater{}bet-\textgreater{}switch-\textgreater{}bet-\textgreater{}show-\textgreater{}prize\\
\textbf{Specific Rules:}\\
In showdown, pick out the players with the lowest combination of cards as the winners.}\\
\textit{\textbf{Badugi}}\\
{\textbf{Suit:} Hearts (H), Diamonds (D), Clubs (C), Spades (S)\\
\textbf{Card Rank:} 1\textless{}2\textless{}3\textless{}4\textless{}5\textless{}6\textless{}7\textless{}8\textless{}9\textless{}10\textless{}11\textless{}12\textless{}13\\
\textbf{Flow:} start-\textgreater{}blind-\textgreater{}deal4-\textgreater{}bet-\textgreater{}switch-\textgreater{}bet-\textgreater{}switch-\textgreater{}bet-\textgreater{}switch-\textgreater{}bet-\textgreater{}show-\textgreater{}prize\\
\textbf{Specific Rules:}\\
In showdown, define a new ranking strategy called ``Badugi''. In given cards, pick out the cards of distinct suits and no pair. If there are more than one cards of the same suit or same value, choose the smaller-ranking one. Hence, the greatest Badugi refers to the the most number of distinct cards and the smallest values.}\\
\textit{\textbf{Badeucey}}\\
{\textbf{Suit:} Hearts (H), Diamonds (D), Clubs (C), Spades (S)\\
\textbf{Card Rank:} 2\textless{}3\textless{}4\textless{}5\textless{}6\textless{}7\textless{}8\textless{}9\textless{}10\textless{}11\textless{}12\textless{}13\textless{}1\\
\textbf{Hand Rank:} High Card\textless{}Pair\textless{}Two Pair\textless{}Three of a Kind\textless{}Full House\textless{}Four of a Kind\\
\textbf{Flow:} start-\textgreater{}blind-\textgreater{}dea5-\textgreater{}bet-\textgreater{}switch-\textgreater{}bet-\textgreater{}switch-\textgreater{}bet-\textgreater{}switch-\textgreater{}bet-\textgreater{}show-\textgreater{}prize\\
\textbf{Specific Rules:}\\
1. In showdown, define a new ranking strategy called ``Badugi''. In given cards, pick out the cards of distinct suits and no pair. If there are more than one cards of the same suit or same value, choose the smaller-ranking one. Hence, the greatest Badugi refers to the the most number of distinct cards and the smallest values.\\2. Pick out the players with the lowest combinations of cards and greatest Badugi as the winners, and split the prize pool equally, and one portion for the winners of the lowest combination, and the other portion for the winners of the greatest Badugi.}\\
\textit{\textbf{Badacey}}\\
{\textbf{Suit:} Hearts (H), Diamonds (D), Clubs (C), Spades (S)\\
\textbf{Card Rank:} 2\textless{}3\textless{}4\textless{}5\textless{}6\textless{}7\textless{}8\textless{}9\textless{}10\textless{}11\textless{}12\textless{}13\textless{}1\\
\textbf{Hand Rank:} High Card\textless{}Pair\textless{}Two Pair\textless{}Three of a Kind\textless{}Full House\textless{}Four of a Kind\\
\textbf{Flow:} start-\textgreater{}blind-\textgreater{}dea5-\textgreater{}bet-\textgreater{}switch-\textgreater{}bet-\textgreater{}switch-\textgreater{}bet-\textgreater{}switch-\textgreater{}bet-\textgreater{}show-\textgreater{}prize\\
\textbf{Specific Rules:}\\
1. In showdown, define a new ranking strategy called ``Badugi''. In given cards, pick out the cards of distinct suits and no pair. If there are more than one cards of the same suit or same value, choose the smaller-ranking one. Hence, the greatest Badugi refers to the the most number of distinct cards and the smallest values.\\2. Pick out the players with the lowest combinations of cards and greatest Badugi as the winners, and split the prize pool equally, and one portion for the winners of the lowest combination, and the other portion for the winners of the greatest Badugi.}\\
\textit{\textbf{2-to-7~single~draw}}\\
{\textbf{Suit:} Hearts (H), Diamonds (D), Clubs (C), Spades (S)\\
\textbf{Card Rank:} 2\textless{}5\textless{}6\textless{}7\textless{}8\textless{}9\textless{}10\textless{}11\textless{}12\textless{}13\textless{}1\\
\textbf{Hand Rank:} High Card\textless{}Pair\textless{}Two Pair\textless{}Three of a Kind\textless{}Straight\textless{}Flush\textless{}Four of a Kind\textless{}Straight Flush\\
\textbf{Flow:} start-\textgreater{}blind-\textgreater{}deal5-\textgreater{}bet-\textgreater{}switch-\textgreater{}bet-\textgreater{}show-\textgreater{}prize\\
\textbf{Specific Rules:}\\
In showdown, pick out the players with the lowest combination of cards as the winners.}
\end{longtblr}

\section{Core Set}
\label{a:core}

\begin{longtblr}[
  caption = {Composition of the core set.},
]{
  width = \linewidth,
  colspec = {Q[200]Q[800]},
  hline{1} = {-}{0.08em},
  hline{2} = {-}{},
}
\small
\textbf{Function}      & \textbf{Instruction}                                                         \\
\texttt{shuffle}		& Generate a deck of all cards and shuffle it following the settings.	 \\
\texttt{blind}		& Set the blind players who are forced to place the bet. The small blind is the one to the left of the button player and the bet is half the minimum bet. The big blind is the one to the left of the small blind and the bet is the minimum bet.	 \\
\texttt{dealx}		& Deal {x} cards to each of the players by order from the top of the deck.	 \\
\texttt{flopx}		& Discard one card, and then flop {x} cards from the top of the deck as the community cards.	 \\
\texttt{switch}		& Discard the specified cards, and then draw the same number of cards from the deck. This is specified by the user: `p{a}: Switch {x}`.	 \\
\texttt{show}		& Show all the hole cards of players and pick out one or more players with the best combination of cards as the winners.	 \\
\texttt{prize}		& Split the prize pool among the winners and recalculate their chips. If more than one players win the game, the pot is split to each of them equally.	 \\
\texttt{show low}		& Show all the hole cards of players and pick out the players with the lowest combination of cards as the winners.	 \\
\texttt{show high}		& Show all the hole cards of players and pick out the players with the highest combination of cards as the winners.	 \\
\texttt{show high low}		& Show all the hole cards of players and pick out the players who have the highest and the lowest combinations of cards as the winners.	 \\
\texttt{prize high low}		& Split the prize pool equally, and one portion for the winners of the highest combination of cards, and the other portion for the winners of the lowest combination of cards.	 \\
\texttt{get straight}		& Pick out the `straight` from given cards, if it exists. If there are more than one, pick out the greater one.	 \\
\texttt{get pair}		& Pick out the `pair` from given cards, if it exists. If there are more than one, pick out the greater one.	 \\
\texttt{get two pair}		& Pick out the `two pair` from given cards, if it exists. If there are more than one, pick out the greater one.	 \\
\texttt{get 3 of a kind}		& Pick out the `three of a kind` from given cards, if it exists. If there are more than one, pick out the greater one.	 \\
\texttt{get 4 of a kind}		& Pick out the `four of a kind` from given cards, if it exists. If there are more than one, pick out the greater one.	 \\
\texttt{get flush}		& Pick out the `flush` from given cards, if it exists. If there are more than one, pick out the greater one.	 \\
\texttt{get full house}		& Pick out the `full house` from given cards, if it exists. If there are more than one, pick out the greater one.	 \\
\texttt{rank low high}		& Rank the given cards from low to high.	 \\
\texttt{rank high low}		& Rank the given cards from high to low.	 \\
\texttt{low suit}		& Pick out the cards of distinct suits. If there are more than one of the same suit, choose the smaller one.	 \\
\texttt{high suit}		& Pick out the cards of distinct suits. If there are more than one of the same suit, choose the greater one.	 \\
\texttt{highest x}		& Choose the top {x} highest-ranking card from given cards.	 \\
\texttt{lowest x}		& Choose the top {x} lowest-ranking cards from given cards.	 \\
\texttt{total bonus}		& Add up all players' chips as the prize pool, which is the total bonus.	 \\
\texttt{bonus for x}		& Average the total bonus or prize pool to {x} winners.	 \\
\texttt{add x chips}		& Add {x} to player {a}'s bet.	 \\
\texttt{drop x chips}		& Take {x} from player {a}'s bet.	 \\
\texttt{bet check}		& Define a user operation called `Check`: Do nothing, only when the bet already matches the highest bet. This is specified by the user: `p{a}: Check`.	 \\
\texttt{bet call}		& Define a user operation called `Call`: Match the amount of the highest bet. This is specified by the user: `p{a}: Call`.	 \\
\texttt{bet raise to x}		& Define a user operation called `Raise`: Increase the bet to a higher bar. This is specified by the user: `p{a}: Raise to {x}`.	 \\
\texttt{bet fold}		& Define a user operation called `Fold`: Discard the hole cards and forfeit all chips already committed to the pot. This is specified by the user: `p{a}: Fold`.	 \\
\texttt{show high x}		& Show all the hole cards of players, and only a combination of {x} hole cards can be considered. Pick out the players with the highest combination of {x} cards as the winners.	 \\
\texttt{show low x}		& Show all the hole cards of players, and only a combination of {x} hole cards can be considered. Pick out the players with the lowest combination of {x} cards as the winners.	 \\
\texttt{highest no pair}		& Select the highest-ranking card with no pair.	 \\
\texttt{lowest no pair}		& Select the lowest-ranking card with no pair.	 \\
\texttt{group suits}		& Group the given cards by their suits.	 \\
\texttt{rank}		& Rank the given cards.	 \\
\texttt{get all}		& List all possible combination of cards from given cards. If there are one more for each combination, choose the greatest one.	 \\
\texttt{len}		& Get the number of cards.
\end{longtblr}

\section{In-Context Learning}
\label{a:icl}
We place three in-context samples for ``All-in'' in the game script. Note that this input is exclusively for samples of \texttt{bet}.

\begin{longtblr}[
  caption = {Few-shot samples for in-context learning.},
]{
  width = \linewidth,
  colspec = {Q[1000]},
  hline{1} = {-}{0.08em},
}
\small
\textbf{\texttt{Game script}}\\
Let's be a Poker engine for Texas hold'em.\\
There are 5 players in the game. The minimum and maximum bet of the game is 2 and 1000.\\
Number: 2$<$3$<$4$<$5$<$6$<$7$<$8$<$9$<$10$<$11$<$12$<$1\\
Suit: H D C S\\
Hand: High Card$<$Pair$<$Two Pair$<$3 of a Kind$<$Flush$<$Full House$<$4 of a Kind$<$Straight Flush\\
Flow: start-$>$blind-$>$deal2-$>$bet-$>$flop3-$>$bet-$>$flop1-$>$bet-$>$flop1-$>$bet-$>$show-$>$prize\\
Define a new player operation in the phase of bet \textasciigrave All-in\textasciigrave. The player puts all his remaining chips into the pot, and is no longer able to make further bet during the game.\\
\\
For example:\\
Input 1:\\
$\vert$chip$\vert$p1: 10/990$\vert$p2: 0/1000$\vert$p3: 0/1000$\vert$p4: 5/995$\vert$p5: 10/990$\vert$p6: 10/990\\
$\vert$message$\vert$engine$\vert$p5$\vert$It's your turn to bet.\\
$\vert$message$\vert$p5$\vert$engine$\vert$All-in.\\
Response 1:\\
$\vert$chip$\vert$p1: 10/990$\vert$p2: 1000/0 (all-in)$\vert$p3: 0/1000$\vert$p4: 5/995$\vert$p5: 1000/0$\vert$p6: 10/990\\
$\vert$message$\vert$engine$\vert$p6$\vert$It's your turn to bet.\\
\\
Input 2:\\
$\vert$chip$\vert$p1: 0/1000$\vert$p2: 0/1000$\vert$p3: 0/1000$\vert$p4: 0/1000$\vert$p5: 0/1000$\vert$p6: 0/1000$\vert$p7: 0/1000\\
$\vert$message$\vert$engine$\vert$p6$\vert$It's your turn to bet.\\
$\vert$message$\vert$p6$\vert$engine$\vert$All-in\\
Response 2:\\
$\vert$chip$\vert$p1: 10/990$\vert$p2: 10/990$\vert$p3: 10/990$\vert$p4: 0/1000$\vert$p5: 10/990$\vert$p6: 1000/0 (all-in)$\vert$p7: 0/1000\\
$\vert$message$\vert$engine$\vert$p7$\vert$It's your turn to bet.\\
\\
Input 3:\\
$\vert$chip$\vert$p1: 24/976$\vert$p2: 30/970$\vert$p3: 0/1000\\
$\vert$message$\vert$engine$\vert$p1$\vert$It's your turn to bet.\\
$\vert$message$\vert$p1$\vert$engine$\vert$All-in\\
Response 3:\\
$\vert$chip$\vert$p1: 1000/0 (all-in)$\vert$p2: 30/970$\vert$p3: 0/1000\\
$\vert$message$\vert$engine$\vert$p2$\vert$It's your turn to bet.\\
\textbf{\texttt{Input}}\\
$\vert$order$\vert$p1$\vert$p2$\vert$p3 (button)$\vert$p4 (small blind)$\vert$p5 (big blind)\\
$\vert$chip$\vert$p1: 10/990$\vert$p2: 0/1000$\vert$p3: 0/1000$\vert$p4: 5/995$\vert$p5: 10/990\\
$\vert$stack$\vert$C12$\vert$S1$\vert$H10$\vert$S2$\vert$S7$\vert$C4$\vert$S5$\vert$C6$\vert$D4$\vert$D12$\vert$C8$\vert$H12$\vert$C11$\vert$D7$\vert$H3$\vert$D3$\vert$D10$\vert$C1$\vert$H5$\vert$H2$\vert$H6...\\
$\vert$hole$\vert$p1$\vert$H8$\vert$D1$\vert$p2$\vert$S6$\vert$C5$\vert$p3$\vert$D8$\vert$S4$\vert$p4$\vert$S3$\vert$C10$\vert$p5$\vert$S12$\vert$D9\\
$\vert$start$\vert$blind$\vert$deal2\\
$\vert$message$\vert$engine$\vert$p2$\vert$It's your turn to bet.\\
$\vert$message$\vert$p2$\vert$engine$\vert$All-in.\\
\textbf{\texttt{Response}}\\
$\vert$order$\vert$p1$\vert$p2$\vert$p3 (button)$\vert$p4 (small blind)$\vert$p5 (big blind)\\
$\vert$chip$\vert$p1: 10/990$\vert$p2: 1000/0 (all-in)$\vert$p3: 0/1000\\
$\vert$stack$\vert$C12$\vert$S1$\vert$H10$\vert$S2$\vert$S7$\vert$C4$\vert$S5$\vert$C6$\vert$D4$\vert$D12$\vert$C8$\vert$H12$\vert$C11$\vert$D7$\vert$H3$\vert$D3$\vert$D10$\vert$C1$\vert$H5$\vert$H2$\vert$H6...\\
$\vert$hole$\vert$p1$\vert$H8$\vert$D1$\vert$p2$\vert$S6$\vert$C5$\vert$p3$\vert$D8$\vert$S4$\vert$p4$\vert$S3$\vert$C10$\vert$p5$\vert$S12$\vert$D9\\
$\vert$start$\vert$blind$\vert$deal2\\
$\vert$message$\vert$engine$\vert$p3$\vert$It's your turn to bet.
\end{longtblr}

\end{document}